\documentclass[letterpaper]{article} % DO NOT CHANGE THIS
\usepackage{aaai2026}  % DO NOT CHANGE THIS
\usepackage{times}  % DO NOT CHANGE THIS
\usepackage{helvet}  % DO NOT CHANGE THIS
\usepackage{courier}  % DO NOT CHANGE THIS
\usepackage[hyphens]{url}  % DO NOT CHANGE THIS
\usepackage{graphicx} % DO NOT CHANGE THIS
\usepackage{natbib}  % DO NOT CHANGE THIS AND DO NOT ADD ANY OPTIONS TO IT
\usepackage{caption} % DO NOT CHANGE THIS AND DO NOT ADD ANY OPTIONS TO IT
\usepackage{algorithm}
\usepackage{algorithmic}

\usepackage{newfloat}
\usepackage{listings}
\DeclareCaptionStyle{ruled}{labelfont=normalfont,labelsep=colon,strut=off} % DO NOT CHANGE THIS
\floatstyle{ruled}
\newfloat{listing}{tb}{lst}{}
\floatname{listing}{Listing}
\usepackage{bm}
\usepackage{booktabs}    
\usepackage{amsmath}
\usepackage{amssymb}
\usepackage{mathtools}
\usepackage{amsthm}

\title{Efficient Diffusion Planning with Temporal Diffusion}
\author {
    Jiaming Guo\textsuperscript{\rm 1}, Rui Zhang\textsuperscript{\rm 1}\thanks{Corresponding Author.}, Zerun Li\textsuperscript{\rm 1,\rm 2,\rm 3}, Yunkai Gao\textsuperscript{\rm 4}, Shaohui Peng\textsuperscript{\rm 5}, Siming Lan\textsuperscript{\rm 4}, Xing Hu\textsuperscript{\rm 1}, Zidong Du\textsuperscript{\rm 1}, Xishan Zhang\textsuperscript{\rm 1,\rm 3}, Ling Li\textsuperscript{\rm 5}
}
\affiliations {
    \textsuperscript{\rm 1}SKL of Processors, Institute of Computing Technology, CAS\\
    \textsuperscript{\rm 2}University of Chinese Academy of Sciences\\
    \textsuperscript{\rm 3}Cambricon Technologies\\
    \textsuperscript{\rm 4}Institute of Al for Industries\\
    \textsuperscript{\rm 5}Intelligent Software Research Center, Institute of Software, CAS\\
   \{guojiaming, zhangrui, lizerun23z\}@ict.ac.cn, ykgao@iaii.ac.cn, pengshaohui@iscas.ac.cn, smlan@iaii.ac.cn, \{huxing,duzidong,zhangxishan\}@ict.ac.cn, liling@iscas.ac.cn
}

\usepackage{bibentry}
\begin{document}

\maketitle

\begin{abstract}
Diffusion planning is a promising method for learning high-performance policies from offline data. To avoid the impact of discrepancies between planning and reality on performance, previous works generate new plans at each time step. However, this incurs significant computational overhead and leads to lower decision frequencies, and frequent plan switching may also affect performance. In contrast, humans might create detailed short-term plans and more general, sometimes vague, long-term plans, and adjust them over time. Inspired by this, we propose the Temporal Diffusion Planner (TDP) which improves decision efficiency by distributing the denoising steps across the time dimension. TDP begins by generating an initial plan that becomes progressively more vague over time. At each subsequent time step, rather than generating an entirely new plan, TDP updates the previous one with a small number of denoising steps. This reduces the average number of denoising steps, improving decision efficiency. Additionally, we introduce an automated replanning mechanism to prevent significant deviations between the plan and reality. Experiments on D4RL show that, compared to previous works that generate new plans every time step, TDP improves the decision-making frequency by 11-24.8 times while achieving higher or comparable performance.
\end{abstract}

\section{Introduction}

Learning effective policies from previously collected sub-optimal data is appealing for reinforcement learning as it circumvents the need for risky or costly environmental interactions \cite{cql,bcq,bear,iql}. Recently, diffusion planning has emerged as a promising approach for this offline reinforcement learning problem \cite{diffuser,dd,videodiff,metadiff}, harnessing the powerful generative power of diffusion models \cite{diffusion} to learn from data and predict future trajectories for effective planning. This method enables decision-making of current action based on long-term plans, thereby mitigating the short-sightedness associated with single-step decision-making.

\begin{figure}[t]
    \centering
    \includegraphics[width=0.9\columnwidth]{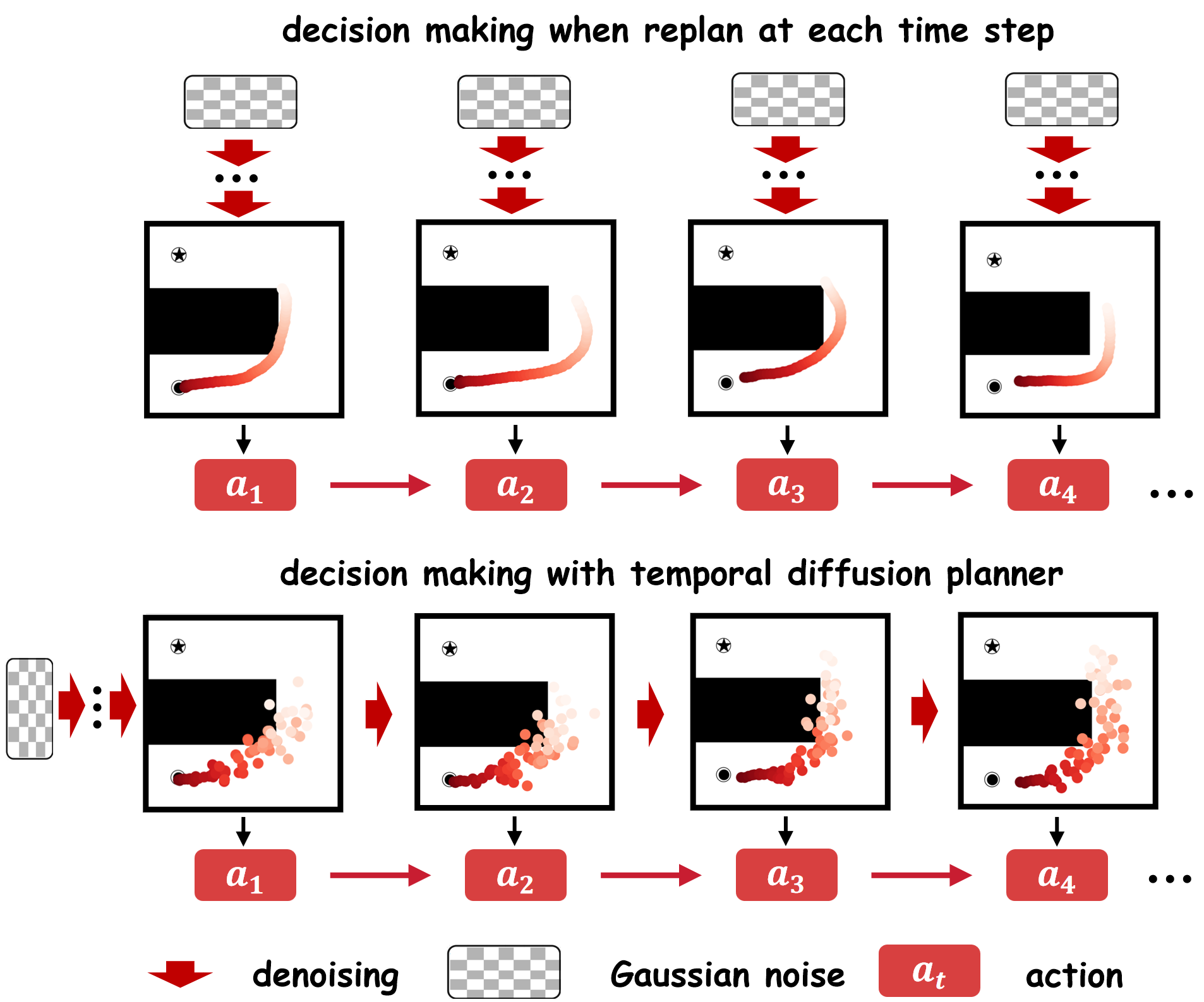}
    %\vspace{-15pt}
    \caption{Comparison of temporal diffusion planner and diffusion planning methods which replan at each time step. The visualization of points in the figure illustrates the x-y coordinates of the planned trajectories.}
    \label{illuer}
    %\vspace{-15pt}
\end{figure}
However, there is often a discrepancy between reality and the plan, and this discrepancy tends to accumulate over time. As a result, rigidly following the original plan without adjustments often leads to poor performance. To mitigate this, previous methods \cite{diffuser,dd} typically execute only the first action of the plan and replan at each time step. However, in diffusion planning, generating a new plan requires multiple denoising steps, involving several forward passes through a large neural network. This leads to expensive computational overhead and low decision frequency. Additionally, frequent plan switching may result in discontinuous actions, which can hinder the agent from achieving a certain goal and degrade the overall performance of the algorithm. Therefore, we explore the possibility of reusing the plan to improve efficiency while reducing the accumulation of planning errors.

How do humans make plans and act according to them? Humans tend to create detailed short-term plans and more general, or even vague, long-term plans. For example, when planning a long journey, a person might detail the first few days' route and stops, while leaving the later part more flexible based on time and weather, focusing only on general destinations. Additionally, people continuously adjust their plans based on the current situation. We can draw inspiration from this to improve the diffusion planning process.

In this paper, we propose the Temporal Diffusion Planner (TDP) for efficient diffusion planning. We present a novel temporal diffusion approach that adjusts the number of diffusion steps along the time horizon. When generating a new plan, TDP creates detailed short-term and vague long-term plans, thereby reducing the cost of generating fully detailed long-term plans. With the temporal diffusion, we can refine the time-evolving plan at each time step by applying a small number of denoising steps to the previous plan. To be specific, at each time step, the execution consumes the beginning of the previous plan while TDP extends the plan to a defined horizon and denoises it based on the current state until the first action is fully detailed. This process corrects the plan based on the current state and allows the plan to be reused and updated over time, as illustrated in Figure \ref{illuer}. To further reduce the accumulation of planning errors, we introduce a mechanism to determine whether to replan based on the degree of discrepancy between the plan and reality. TDP maximizes the reuse of previously generated plans to reduce computational load, allowing for efficient diffusion planning, while the plan refinement and automated replanning ensure consistent performance.

In our experiments, we validate the proposed Temporal Diffusion Planner using the D4RL \cite{d4rl} benchmark. The results demonstrate that the Temporal Diffusion Planner achieves 11 to 24.8 times greater decision-making efficiency compared to methods that replan at each step while maintaining comparable or improved performance.

\section{Related Work}

\paragraph{Offline Reinforcement Learning.}
Offline reinforcement learning (RL) focuses on learning policies from static datasets. A series of works have been developed based on off-policy RL methods. These approaches have to mitigate estimation errors for out-of-distribution actions arising from distributional dissimilarities by constraining the learned policy to remain closer to the behavior policy \cite{bear,brac,awr,spot}, penalize learned value functions to produce more conservative estimates \cite{nachum2019algaedice,bcq,cql,mcq}, or performing a single step of evaluation followed by a single policy improvement step rather than iterative policy evaluation \cite{brandfonbrener2021offline,iql}.  Some works categorized as model-based offline RL involve a model for data generation \cite{morel,argenson2020model,yu2020mopo, yang2021pareto} or trajectory search \cite{tt}. There are also return conditioned approaches that learn policies via return conditioned behavioral cloning from the dataset \cite{dt,rvs}. In this work, we study using the diffusion model for trajectory generation and planning in decision-making tasks.

\paragraph{Diffusion Models for Decision Making.}
The denoising diffusion probabilistic model (Diffusion) \cite{diffusion} frames the generation process as a Markov decision process (MDP) tied to noise, comprising a forward process that adds noise and a reverse process that denoises to recover the original distribution.
Leveraging its strong capacity to model arbitrary distributions, Diffusion models have gained increasing traction in decision-making tasks.
Several works model policies with diffusion models to capture multimodal action distributions \cite{diffusionql,edp,dp}.
Others utilize diffusion models for reinforcement learning data generation \cite{dedm,ser}.
Recent research also explores diffusion models for trajectory generation and planning, a category our work falls into.
Diffuser \cite{diffuser} employs a diffusion model with classifier-guided sampling to generate planning trajectories, while DD \cite{dd} designs a return-conditioned diffusion model for trajectory generation and uses an inverse dynamic model for action prediction.
Building on these pioneering methods, diffusion-based planning has been applied to diverse domains \cite{metadiff,force,madiff,rgg} to achieve higher performance.
Several studies \cite{hdmi,hd,diffuerlite} turn to employ multiple diffusion models to form a hierarchical frameworks and make decisions at multiple level, while we focus on single-level planning which is the fundamental component and study improving the denoising manner to achieve efficient diffusion planning.
With the same purpose of us, prior methods add noise to previous plans as a warm start for subsequent planning iterations to reduce the denoising steps \cite{diffuser,tree}.
However, they trade sample quality to achieve acceleration and the resulting speedup remains limited under constraints on performance degradation,
In contrast, our approach significantly boosts efficiency while preserving performance.

% The denoising diffusion probabilistic model (Diffusion) \cite{diffusion} formulates generation as an MDP involving noise, with a forward process adding noise and a reverse process denoising to recover the original distribution. Due to its ability to model complex distributions, Diffusion has been widely adopted in decision-making.
% Diffusion-QL \cite{diffusionql} models the policy via diffusion and integrates a TD3-BC-style loss using behavior cloning as a constraint. Diffuser \cite{diffuser} uses classifier-guided sampling for planning, while Decision Diffuser (DD) \cite{dd} introduces return-conditioning and inverse dynamics for action inference. These approaches have inspired diffusion-based planning across diverse domains \cite{videodiff,metadiff,adaptdiffuser,motiondiffuser,refinediffuser,multitaskdiffuser,force,madiff,rgg}.
% To improve efficiency, some methods \cite{diffuser,tree} warm-start planning by adding noise to prior trajectories, reducing denoising steps at the cost of sample quality and limited speedup. Others \cite{hdmi,hd,ld,diffuerlite} adopt hierarchical planning to enhance performance. In contrast, we focus on single-level planning and propose improving the denoising process for more efficient diffusion planning.

\begin{figure*}[t]
    \centering
    \includegraphics[width=0.92\textwidth]{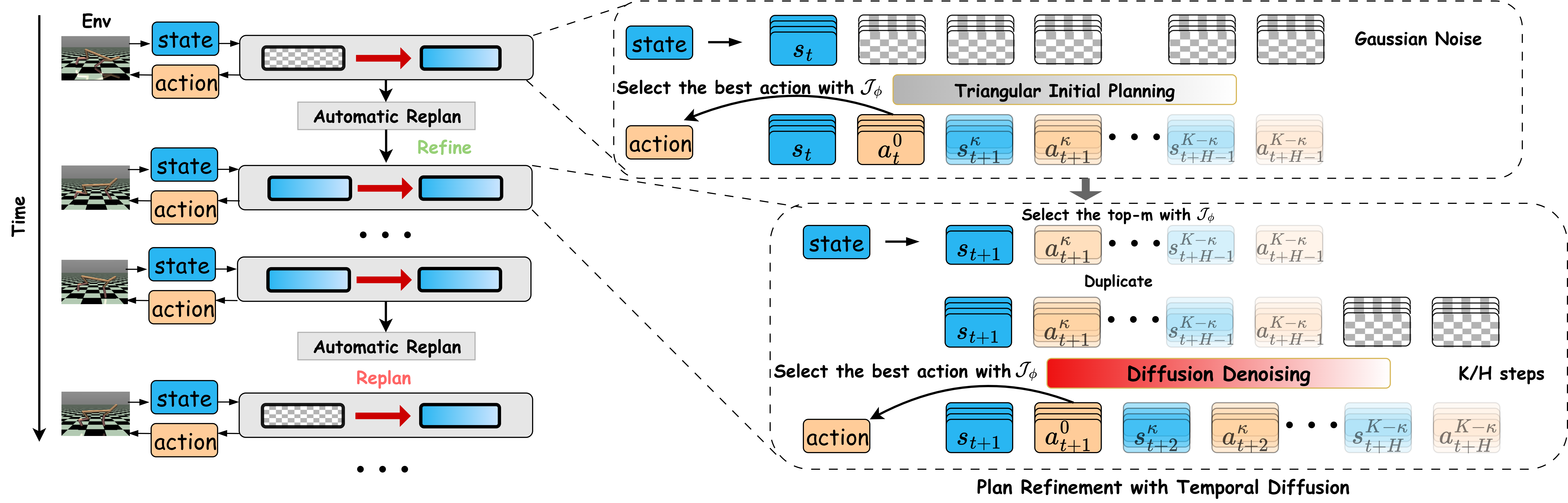}
    %\vspace{-8pt}
    \caption{Overview of the Temporal Diffusion Planner. TDP starts with Triangular Initial Planning, creating detailed short-term and vague long-term plans. It then refines previous time-evolving plan at each step via temporal diffusion, applying very small number of denoising steps until automatic-replanning is triggered. }
    \label{tdiff}
    %\vspace{-10pt}
\end{figure*}
\section{Preliminaries}

\paragraph{Problem Setup.}
We consider the sequential decision-making problem as a discounted Markov Decision Process, defined as the 6-tuple {\small$\langle \mathcal{S},\mathcal{A},\mathcal{P},\rho_0,r,\gamma \rangle$}, where {\small$\mathcal{S} \subseteq \mathbb{R}^m$} and {\small$\mathcal{A} \subseteq \mathbb{R}^n$} are the state and action spaces, {\small$\mathcal{P}$} is the transition distribution, {\small$\rho_0$} the initial state distribution, {\small$r$} the reward function, and {\small$\gamma \in (0,1)$} the discount factor.
The agent acts with a stochastic policy {\small $\pi: \mathcal{S} \rightarrow  \mathcal{A}$} and generates a trajectory sequence {\small$\tau = (s_0,a_0,\dots)$} following {\small$s_0 \sim \rho_0, a_t \sim \pi(a_t|s_t), s_{t+1} \sim \mathcal{P}(s_{t+1}|s_t,a_t)$}. The standard objective
in RL is to optimize the expected return {\small$\eta(\pi)= \mathbb{E}_\tau[\sum^\infty_{t=0}\gamma^t r(s_t,a_t)]$}. We define the value function as {\small$V(s_t) = \mathbb{E}_{a_t,s_{t+1},\dots}[\sum_{l=0}^\infty \gamma^l r(s_{t+l},a_{t+l})|s_t]$}. We define the expected return achieved under the optimal policy as {\small$Q^*(s,a)=\mathbb{E}_{\tau \sim p_{\pi^*}}[\sum^\infty_{t=0}\gamma^t r(s_t,a_t)|s_0=s,a_0=a]$}. 

In offline RL, agents focus on learning from a fixed dataset {\small $D=\{(s, a,r,s')\}$} pre-collected using an unknown behavior policy {\small$\pi_{\beta}(a|s)$}. 
To solving this problem, diffusion planning methods employ a diffusion model to generate a trajectory {\small$\tau = \{s_t,a_t,s_{t+1},a_{t+1},\dots,s_{t+H-1},a_{t+H-1}\}$} or {\small$\tau = \{s_t,s_{t+1},\dots,s_{t+H-1}\}$}, where {\small $H$} is the planning horizon. The agent will interact with the environment according to the trajectory, e.g. directly using the generated first action {\small$a_t$} \cite {diffuser} or predicting the action with an inverse dynamic model {\small$p(a_t|s_t,s_{t+1})$ }\cite{dd}. The trajectory should maximize the object value {\small$\mathcal{J}(\tau)$}, typically defined as expected return in offline RL.

% \subsection{Diffusion Probabilistic Model}
\paragraph{Diffusion Probabilistic Model.} Diffusion probabilistic models \cite{diffusion} are a class of latent variable generative models that learn the data distribution {\small$q(\bm{x})$} from a dataset. 
Diffusion models contain a forward noising process for data generation and a trainable reverse denoising process.
In the forward process, Gaussian noise is added to origin data {\small$x_{0}$} over {\small$K$} steps, following a variance schedule {\small$\beta_{1:K}$, $0<\beta_{1}<\beta_{2}<\cdots<\beta_{K}<1$}, generating a sequence of {\small$\bm{x}_{1:K}$} until it nearly becomes pure Gaussian noise. Then the relation between {\small$\bm{x}_{k-1}$ and $\bm{x}_{k}$} is {\small$\bm{x}_{k}=\sqrt{1-\beta_{k}}\bm{x}_{k-1}+\sqrt{\beta_{k}}z_{k}$}, where {\small$z_{k}\sim\mathcal{N}(0,I)$}.
The relation between {\small$\bm{x}_{0}$ and $\bm{x}_{k}$} is {\small$\bm{x}_{k}=\sqrt{\bar{\alpha}_{k}}\bm{x}_{0}+\sqrt{1-\bar{\alpha}_{k}}z_{k}$}, where {\small$\alpha_{k}=1-\beta_{k}$ and $\bar{\alpha}_{k}=\alpha_{1}\alpha_{2}\dots\alpha_{k}$}.
The reverse denoising process is constructed as {\small$p(\bm{x}_{0:K})\sim\mathcal{N}(\bm{x}_{k};0,I)\prod_{k=1}^{K}p(\bm{x}_{k-1}|\bm{x}_{k})$}. Through the Bayesian equation, 
{\small
$$
        p(\bm{x}_{k-1}|\bm{x}_{k}):=\mathcal{N}(\bm{x}_{k-1}|\mu_\theta(\bm{x}_{k},k);\Sigma_{k}) $$
        $$
        \mu_\theta(\bm{x}_{k},k)=\frac{1}{\sqrt{\alpha_{k}}}(\bm{x}_{k}-\frac{1-\alpha_{k}}{\sqrt{1-\bar{\alpha}_{k}}}\bm{\epsilon}_{\theta}), \Sigma_{k}=\frac{1-\bar{\alpha}_{k-1}}{1-\bar{\alpha}_{k}}\beta_{k}.$$
}
Thus, the reverse denoising process $p$ is parameterized through the noise model $\epsilon_{\theta}$, and the optimization object is to maximize the evidence lower bound, the corresponding loss is 
{\small
$L_{\theta}=\mathbb{E}_{\bm{x}_{0},\bm{\epsilon}}[\Vert\bm{\epsilon}-\bm{\epsilon}_{\theta}(\sqrt{\bar{\alpha}_{k}}\bm{x}_{0}+\sqrt{1-\bar{\alpha}_{k}}\bm{\epsilon}, k)]\Vert^{2}.$
}
\paragraph{Conditional Diffusion Sampling.} To solve reinforcement learning problems with diffusion planning, the generated trajectory should maximize the object function {\small$\mathcal{J}(\tau)$}, which is a conditional diffusion sampling problem. There are two main approaches for conditional diffusion sampling: classifier-guided \cite{DBLP:conf/icml/NicholD21} and classifier-free \cite{DBLP:journals/corr/abs-2207-12598}. 
Following Diffuser, we use the classifier-guided method. This method first trains a separate model {\small$\mathcal{J}_\phi$} as the objective function. Then it uses the gradients of {\small$\mathcal{J}_\phi$}  to guide the trajectory sampling procedure by modifying the means $\mu$ of the reverse process:
{\small$p_\theta(\bm{x}_{k-1}|\bm{x}_{k}):=\mathcal{N}(\bm{x}_{k-1}|\mu+\alpha\Sigma_{k}\nabla_{\mu}\mathcal{J}(\mu);\Sigma_{k}),$}
where {\small$\mu$} refers to {\small$\mu_\theta(\bm{x}_{k},k)$} and {\small$\alpha$} is the scale.

\section{Method}

% This section introduces the components of the Temporal Diffusion Planner. We begin by describing the process that generates the initial plan triangularly. Then we introduce the temporal diffusion process which refines the plan along the time axis. Finally, we describe how to automatically replan to avoid performance degradation.

In this section, we first describe the proposed temporal diffusion model, along with its training process and architecture. We then introduce the Temporal Diffusion Planner (TDP), which consists of three main components: 1) the process for generating the initial plan in a triangular manner, 2) the process for refining the plan over time using temporal diffusion, and 3) the mechanism for automatic replanning to prevent performance degradation.

\subsection{Temporal Diffusion}
The standard diffusion planning paradigm generates a clear plan of horizon $H$ at each decision-making step, or in other words, predicts all timesteps of a plan simultaneously. However, as the poem line goes, "\textit{The best laid plans of mice and men often go awry}", executing a complete plan over time is difficult due to cumulative forecasting errors or unexpected environment dynamics. Thus, the further forward the plan, the less important its details are to the current decision-making. When humans plan, they tend to make detailed plans for the near future and more general, macro plans for the long term. Inspired by this, we adopt a similar strategy when using diffusion to generate plans.

We propose the temporal diffusion which uses different diffusion steps at different time steps of the plan. 
Specifically, for a trajectory of length {\small$H$: $\tau = (s_t,a_t,s_{t+1},a_{t+1},\dots,s_{t+H-1},a_{t+H-1})$}, we define the noise array of the trajectory as {\small$\bm{x}_k(\tau)$}, where {\small$k$} is the diffusion step for the start of the noise array. For each step in time, the corresponding diffusion step increases by {\small$\kappa$}, where  {\small$\kappa \in \mathbb{N_+}$}  and {\small$\kappa \leq K/H$}. In other words, during the denoising process, as time progresses, the number of denoising steps decreases by $\kappa$ at each step. 
In this way, the noise array of the plan gets more and more ambiguous over time.
We can formulate the noise array {\small$\bm{x}_k(\tau)$} as:
{\small$\bm{x}_k(\tau)=\{\bm{x}_{k+i\kappa}(s_{t+i},a_{t+i})\}_{i=0}^{i=H-1}$},
where {\small$\bm{x}_k(s,a)=(\sqrt{\bar{\alpha}_{k}}s+\sqrt{1-\bar{\alpha}_{k}}\bm{\epsilon},\sqrt{\bar{\alpha}_{k}}a+\sqrt{1-\bar{\alpha}_{k}}\bm{\epsilon})$}. For brevity,
we use {\small$s_t^k$} and {\small$a_t^k$} to represent {\small$\sqrt{\bar{\alpha}_{k}}s_t+\sqrt{1-\bar{\alpha}_{k}}\bm{\epsilon}$} and {\small$\sqrt{\bar{\alpha}_{k}}a_t+\sqrt{1-\bar{\alpha}_{k}}\bm{\epsilon}$} respectively in the following. Note that we cut off the noise array when the diffusion step {\small$k+i\kappa$} is greater than or equal to the maximum number of diffusion steps {\small$K$}.
Thus with different diffusion steps {\small$k$}, the length of the noise array {\small$\bm{x}_k(\tau)$} may be different. 

For conditional sampling, we train a separate model {\small$\mathcal{J}_\phi(\bm{x}_k(\tau))$} to predict the value function {\small$V(\tau)$} conditioned on the original trajectory with the noise array as input. Then the reverse denoising process transitions can be formulated as:
{\small
\begin{equation}
\label{denoise}
    p_\theta(\bm{x}_{k-1}(\tau)|\bm{x}_k(\tau)):=\mathcal{N}(\bm{x}_{k-1}(\tau)|\mu+\alpha\Sigma_{k}\nabla_{\mu}\mathcal{J}(\mu);\Sigma_{k}),
\end{equation}
}
where {\small$\mu$} refers to {\small$\mu_\theta(\bm{x}_{k}(\tau),k)$} and {\small$\alpha$} is the scale. When the length of {\small$\bm{x}_{k-1}(\tau)$} is larger than {\small$\bm{x}_k(\tau)$}, we can pad {\small$\bm{x}_k(\tau)$} with noise array sampled from {\small$\mathcal{N}(\bm{0},\bm{I})$}. Besides, we always replace the first state of the noise array with the current state {\small$s_t$} during training and inference. In this way, the diffusion model denoises the array conditioned on the current state.

\paragraph{Training Process.} 
For the Temporal Diffusion, we train the reverse denoising process {\small$p_{\theta}$}, parameterized through the noise model {\small$\epsilon_{\theta}$} as traditional diffusion models \cite{diffusion,diffuser,dd}. 
The key difference lies in the diffusion process, where the diffusion steps differ across time steps. As a result, we adopt a temporal noise addition strategy in the forward process of diffusion. Specifically, for a trajectory 
{\small$\tau = (s_t,a_t,s_{t+1},a_{t+1},\dots,s_{t+H-1},a_{t+H-1})$}, we sample noise {\small$\epsilon \sim \mathcal{N}(\bm{0},\bm{I})$} and the diffusion step {\small$k \sim \mathcal{U}\{1,\dots,K\}$} to obtain a noise array of the trajectory {\small$\bm{x}_k(\tau)$}. 
Then we train the noise model with the following loss:
\begin{equation}
\small
   \mathcal{L}(\theta)=\mathbb{E}_{\tau \in D, k}[\Vert \epsilon-\epsilon_\theta(\bm{x}_k(\tau),k)\Vert^2]  
\end{equation}

\paragraph{Architecture.}
% We parameterize $\epsilon_\theta$ with a temporal U-Net following Diffuser \cite{diffuser}. This neural network consists of repeated convolutional residual blocks and adopts an architecture similar to U-Nets in image-based diffusion models, substituting two-dimensional spatial convolutions with one-dimensional temporal convolutions. As a fully convolutional model, the prediction horizon is determined by the input dimension rather than the model architecture. This design aligns with the requirement of temporal diffusion for handling variable input and output lengths.
We parameterize $\epsilon_\theta$ with a temporal U-Net \cite{diffuser}, comprising repeated convolutional residual blocks. It resembles image-based U-Nets but replaces 2D spatial convolutions with 1D temporal ones. As a fully convolutional model, the prediction horizon depends on input dimension, which aligns with the requirement of temporal diffusion for handling variable data lengths.

%现在我们介绍如何用temporal diffusion model来产生计划。由于temporal diffusion在不同时间步的diffusion step是不同的，从头开始生成计划的过程像是一个三角形，因此我们称这个过程为Triangular Initial Planning。
\begin{table*}[tb]
\setlength{\tabcolsep}{1mm}
\centering
{\small
\begin{tabular}{@{}llcccccccccc@{}}
\toprule
\textbf{Dataset} & \multicolumn{1}{r}{\textbf{Environment}} & \textbf{BC}    & \textbf{CQL}   & \textbf{IQL}   & \textbf{DT}    & \textbf{TT}    & \textbf{MOReL} & \textbf{Diffuser} & \textbf{DD}   & \textbf{TDP}        & \textbf{TDP\textsubscript{Inv}}                          \\ \midrule
Medium-Expert    & HalfCheetah                              & 55.2           & 91.6           & 86.7           & 86.8           & \textbf{95.0}  & 53.3           & 88.9              & \textbf{90.6} & \textbf{91.6} \scriptsize{\raisebox{1pt}{$\pm 0.26$}}  & \textbf{93.9} \scriptsize{\raisebox{1pt}{$\pm 0.2$}} \\
Medium-Expert    & Hopper                                   & 52.5           & 105.4          & 91.5           & \textbf{107.6} & \textbf{110.0} & \textbf{108.7} & 103.3             & \textbf{111.8}                    & \textbf{112.4} \scriptsize{\raisebox{1pt}{$\pm 0.17$}} & \textbf{110.1} \scriptsize{\raisebox{1pt}{$\pm 0.92$}}                   \\
Medium-Expert    & Walker2d                                 & \textbf{107.5} & \textbf{108.8} & \textbf{109.6} & \textbf{108.1} & 101.9          & 95.6           & \textbf{106.9}    & \textbf{108.8}                    & \textbf{108.2} \scriptsize{\raisebox{1pt}{$\pm 0.19$}} & \textbf{108.4 } \scriptsize{\raisebox{1pt}{$\pm 0.07$}}                    \\ \midrule
Medium           & HalfCheetah                              & 42.6           & 44.0           & \textbf{47.4}  & 42.6           & \textbf{46.9}  & 42.1           & 42.8              & \textbf{49.1}                     & 43.9 \scriptsize{\raisebox{1pt}{$\pm 0.24$}}           & 43.6 \scriptsize{\raisebox{1pt}{$\pm 0.19$}}                             \\
Medium           & Hopper                                   & 52.9           & 58.5           & 66.3           & 67.6           & 61.1           & \textbf{95.4}  & 74.3              & 79.3                              & 74.0 \scriptsize{\raisebox{1pt}{$\pm 3.41$}}           & \textbf{96.3} \scriptsize{\raisebox{1pt}{$\pm 1.17$}}                     \\
Medium           & Walker2d                                 & 75.3           & 72.5           & 78.3           & 74.0           & \textbf{79.0}  & 77.8           & \textbf{79.6}     & \textbf{82.5}                     & \textbf{78.8} \scriptsize{\raisebox{1pt}{$\pm 1.12$}}  & \textbf{80.1} \scriptsize{\raisebox{1pt}{$\pm 1.22$}}                     \\ \midrule
Medium-Replay    & HalfCheetah                              & 36.6           & \textbf{45.5}  & \textbf{44.2}  & 36.6           & 41.9           & 40.2           & 37.7              & 39.3                              & 39.2 \scriptsize{\raisebox{1pt}{$\pm 0.32$}}           & 39.7 \scriptsize{\raisebox{1pt}{$\pm 0.28$}}                              \\
Medium-Replay    & Hopper                                   & 18.1           & \textbf{95.0}  & 94.7           & 82.7           & 91.5           & 93.6           & 93.6              & \textbf{100}                      & 94.9 \scriptsize{\raisebox{1pt}{$\pm 0.64$}}           & \textbf{96.1} \scriptsize{\raisebox{1pt}{$\pm 1.50$}}                    \\
Medium-Replay    & Walker2d                                 & 26.0           & 77.2           & 73.9           & 66.6           & 82.6           & 49.8           & 70.6              & 75                                & 71.2 \scriptsize{\raisebox{1pt}{$\pm 2.83$}}           & 70.5 \scriptsize{\raisebox{1pt}{$\pm 6.15$}}                              \\ \midrule
\multicolumn{2}{c}{\textbf{Average}}                        & 51.9           & 77.6           & 77.0           & 74.7           & 78.9           & 72.9           & 77.5              & \textbf{81.8}                     & \textbf{79.4}       & \textbf{82.1}                              \\ \midrule
Mixed            & Kitchen                                  & 51.5           & 52.4           & 51             & -              & -              & -              & 50.0              & \textbf{65}                       & 57.5                & 57.5                                       \\
Partial          & Kitchen                                  & 38             & 50.1           & 46.3           & -              & -              & -              & 56.2              & 57                                & \textbf{65.0}       & 60.0                                       \\ \midrule
\multicolumn{2}{c}{\textbf{Average}}                        & 44.8           & 51.2           & 48.7           & -              & -              & -              & 53.1              & \textbf{61.0}                     & \textbf{61.3}       & \textbf{58.8}                              \\ \bottomrule
\end{tabular}%
}
\caption{\textbf{Offline Reinforcement Learning Performance.} The performance of TDP, TDP\textsubscript{Inv} and a variety of prior algorithms on the D4RL tasks \cite{d4rl}. Results for TDP and TDP\textsubscript{Inv} correspond to the mean and standard error over 5 random seeds. Following Diffuser \cite{diffuser}, we emphasize in bold scores within 5 percent of the maximum per task ($\geq 0.95 \cdot max$)}
\label{main_results}
%\vspace{-5pt}
\end{table*}

\begin{table}[tb]
\setlength{\tabcolsep}{1mm}
\centering
{\small
\begin{tabular}{@{}lccccc@{}}
\toprule
\textbf{Environment} & \textbf{Metric}                                                      & \textbf{Diffuser}                                   & \textbf{DD}                                        & \textbf{TDP}                                                  & \textbf{TDP\textsubscript{Inv}}                                         \\ \midrule
Mujoco               & \begin{tabular}[c]{@{}c@{}}Runtime (s)\\ Frequency (Hz)\end{tabular} & \begin{tabular}[c]{@{}c@{}}0.734\\ 1.4\end{tabular} & \begin{tabular}[c]{@{}c@{}}1.65\\ 0.6\end{tabular} & \textbf{\begin{tabular}[c]{@{}c@{}}0.086\\ 11.6\end{tabular}} & \begin{tabular}[c]{@{}c@{}}0.125\\ 8.0\end{tabular}  \\ \midrule
Kitchen              & \begin{tabular}[c]{@{}c@{}}Runtime (s)\\ Frequency (Hz)\end{tabular} & \begin{tabular}[c]{@{}c@{}}0.731\\ 1.4\end{tabular} & \begin{tabular}[c]{@{}c@{}}1.57\\ 0.64\end{tabular}  & \textbf{\begin{tabular}[c]{@{}c@{}}0.052\\ 19.2\end{tabular}} & \begin{tabular}[c]{@{}c@{}}0.063\\ 15.9\end{tabular} \\ \midrule
\multicolumn{2}{c}{\textbf{Average}}                                                        & \begin{tabular}[c]{@{}c@{}}0.733\\ 1.4\end{tabular} & \begin{tabular}[c]{@{}c@{}}1.61\\ 0.62\end{tabular} & \textbf{\begin{tabular}[c]{@{}c@{}}0.069\\ 15.4\end{tabular}} & \begin{tabular}[c]{@{}c@{}}0.094\\ 12.0\end{tabular} \\ \bottomrule
\end{tabular}%
}
\caption{\textbf{Efficiency comparison of diffusion planning methods.} The runtime cost per step of decision-making and the decision frequency.}
\label{efficiency}
%\vspace{-10pt}
\end{table}
\begin{figure}[t]
    \centering
    \includegraphics[width=0.9\columnwidth]{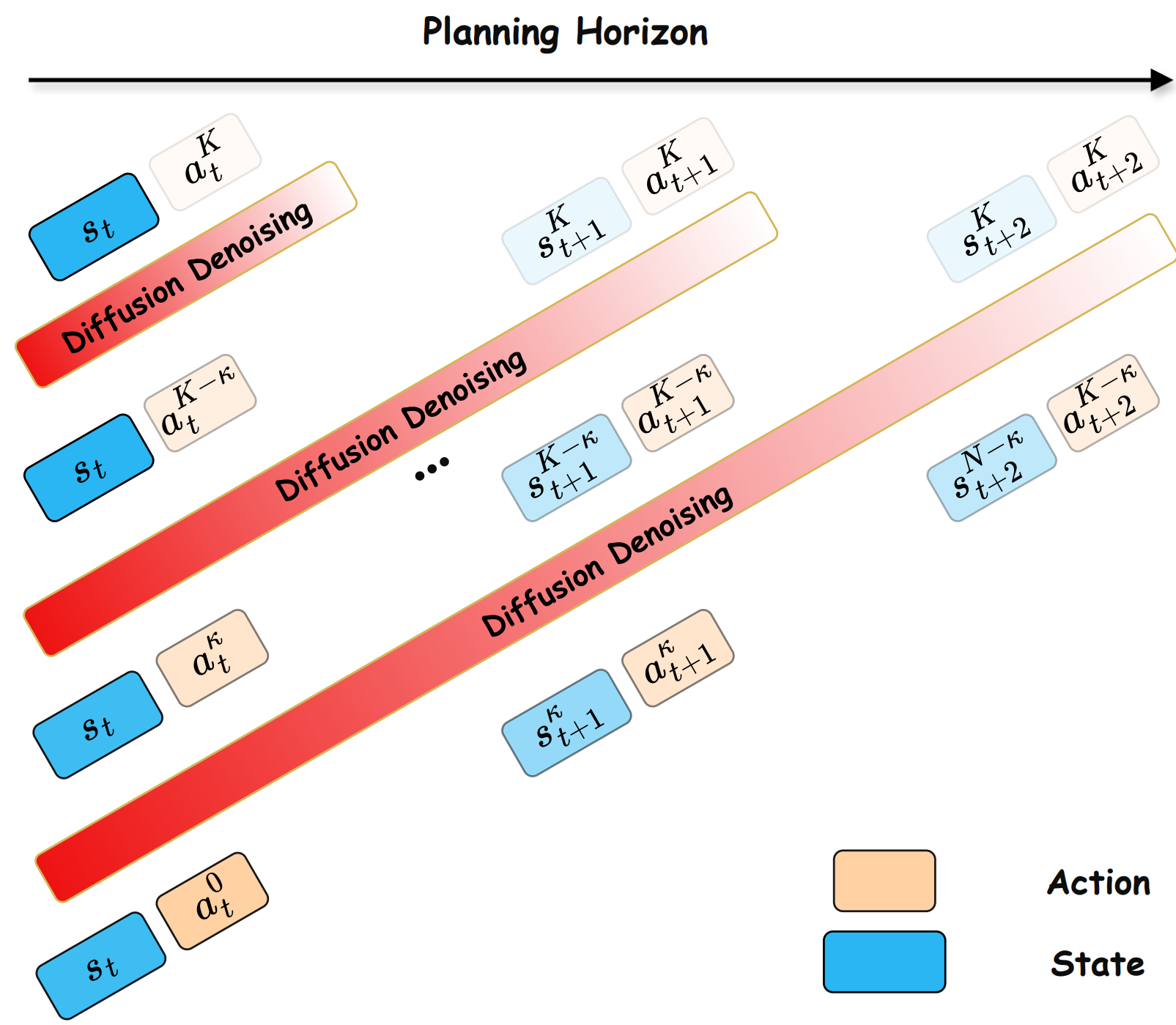}
    %\vspace{-10pt}
    \caption{Triangular Initial Planning when the planning horizon is set as 3. $s_{t+i}^K, a_{t+i}^K$ are sampled from Gaussian noise.}
    \label{triangular}
    %\vspace{-12pt}
\end{figure}
\subsection{Triangular Initial Planning}
We now explain how to generate a plan using the temporal diffusion model. In practice, we set {\small$\kappa$} as {\small$K/H$} and assume {\small$H$} can divide {\small$K$} exactly.
Since the diffusion steps vary across time steps, the process of generating a plan from scratch resembles a triangular shape, with the slope of the hypotenuse denoted as {\small$\kappa$}. Thus, we refer to this process as Triangular Initial Planning. We illustrate the process when the planning horizon is set as 3 in Figure \ref{triangular}. We start by sampling the first state-action pair {\small$\{s_t^K,a_t^K\}$} from {\small$\mathcal{N}(\bm{0},\bm{I})$} and replace {\small$s_t^K$} with the current state {\small$s_t$}. Then we use the diffusion model for denoising {\small$\kappa$} steps according to Equation \ref{denoise} and also replace the first state to obtain {\small$(s_t,a_t^{K-\kappa})$}. Next, we sample the second state-action pair {\small$\{s_{t+1}^K,a_{t+1}^K\}$} from {\small$\mathcal{N}(\bm{0},\bm{I})$} and append to the previous state-action pair:  {\small$\{s_t,a_t^{K-\kappa},s_{t+1}^K,a_{t+1}^K\}$}. As the diffusion model is able to accept variable length inputs, we use it to denoise the new sequence {\small$\kappa$} steps:  {\small$\{s_t^{K-2\kappa},a_t^{K-2\kappa},s_{t+1}^{K-\kappa},a_{t+1}^{K-\kappa}\}$}. And so on, we repeat the sampling and denoising process to finally get the initial plan: {\small$\{s_t,a_t^{0},s_{t+1}^{\kappa},a_{t+1}^{\kappa},\cdots,s_{t+H-1}^{K-\kappa},a_{t+H-1}^{K-\kappa}\}$} which can also be represented as {\small$\bm{x}_0(\tau)$}. In this way, we get plans that blur over time and at the same time reduce the amount of computation by about half. In practice,   
we use this process simultaneously to obtain {\small$M$} candidate plans {\small$\{\bm{x_0}(\tau)\}_M$}. We predict the cumulative rewards of the candidate plans with {\small$\mathcal{J}_\phi$} and select the best plan to extract the action.

\subsection{Plan Refinement with Temporal Diffusion}

In the proposed TDP, we do not replan at each step as in the previous works but rather use the temporal diffusion model to refine the initial plan at each time step according to the new state. This will greatly reduce the average diffusion steps and improve efficiency. We illustrate the refinement process in Figure \ref{tdiff} and provide the details below.

In the initial plan, only the action of the first step is totally \textit{clear} as it is obtained from the full {\small$K$} denoising steps. After we execute the first action {\small$a_t^0$} and remove {\small$s_t,a_t^0$}, the remaining candidate plans are {\small$\{s_{t+1}^{\kappa},a_{t+1}^{\kappa},\ldots,s_{t+H-1}^{K-\kappa},a_{t+H-1}^{K-\kappa}\}_M$}, which is exactly {\small$\{\bm{x}_{\kappa}(\tau_{t+1})\}_M$}, where {\small$\tau_{t+1}$} means the trajectory starts from {\small$s_{t+1}$}. Thus, we can naturally use the temporal diffusion model to refine the plan by denoise the remaining candidate plans $\kappa$ steps. In practice, we do this with a candidate selection process.
We select the top-m plans from these candidate plans based on the objective function {\small$\mathcal{J}_\phi$} to eliminate inferior plans while maintaining diversity. We replace {\small$s_{t+1}^{\kappa}$} with the current state {\small$s_{t+1}$} and duplicate the plans to obtain {\small$M$} candidate plans. Then we sample {\small$M$} state-action pair {\small$\{s_{t+H}^K,a_{t+H}^K\}$} from Gaussian Noise {\small$\mathcal{N}(\bm{0},\bm{I})$} and append to the candidate plans. Finally, we diffuse these candidate plans {\small$\kappa$} steps and obtain the refined plans: 
{\small$\{\bm{x}_0(\tau_{t+1})\}_M=\{s_{t+1},a_{t+1}^{0},s_{t+2}^{\kappa},a_{t+2}^{\kappa},\cdots,s_{t+H}^{K-\kappa},a_{t+H}^{K-\kappa}\}_M$}.
As {\small$\kappa$} is a small number, e.g. usually set as 1 in our experiments, the process is very efficient.
Same as the triangular initial planning process, we also execute the first action of the plan with the highest objective function {\small$\mathcal{J}_\phi$}. For each new time step, we can repeat this process to refine the plan from the previous step to obtain the new plan and the current action. Note that the horizon of the initial plan does not limit the amount of refinement, because each time step we append a new state-action pair at the end of the remaining plans thus obtaining a full plan of length {\small$H$}.

\subsection{Automatic Replanning}
Although plan refinement can adjust the plan based on the new state, the plan may still deviate more and more over time, leading to the refinement process failing to guarantee a good plan. To address this issue, we propose to replan automatically based on the plan errors and design two criteria: 

\paragraph{State-based criterion.}
We can decide whether to replan based on how far the current state obtained from the environment differs from the current state in the previous plan. In the plan obtained in the previous time step, the current state {\small$s_{t+1}^{\kappa}$} has not yet gone through the complete diffusion process. According to the diffusion forward noising process, we can calculate the probability distribution of {\small$s_{t+1}^{*\kappa}$} after adding noise over $\kappa$ steps to the ground truth {\small$s_{t+1}^*$: $p(s_{t+1}^{*\kappa}|s_{t+1}^*)$}. In this way, we can use the probability of {\small$s_{t+1}^{\kappa}$} in this distribution as the criterion of replanning:
{\small
$$\mathcal{C}_{state}= -log p(s_{t+1}^\kappa|s_{t+1}^*).$$
}
\paragraph{Value-based criterion.}
When we generate the plan or refine the plan, we estimate the objective function {\small$\mathcal{J}_\phi$} for the plan at the same time. The objective function {\small$\mathcal{J}_\phi(\bm{x}_0(\tau))$} estimates the value function conditioned on the plan: $V(\tau)$. We use the TD-loss for values for new plan {\small$\tau_{t+1}$} and the previous plan {\small$\tau$ }as the criterion of replanning:
{\small
$$\mathcal{C}_{value}= \mathcal{J}_\phi(\bm{x}_0(\tau))-r-\gamma\mathcal{J}_\phi(\bm{x}_0(\tau_{t+1})).$$
}
At every time step, we compute the criterion and if the criterion is greater than a threshold, we use the Triangular Initial Planning process to get the plan. In practice, we use the value-based criterion and we show how the design choice of the criterion affects the performance in Ablation Study. We provide the algorithm of the TDP in the Appendix.

% \begin{algorithm}[tb]
%    \caption{Bubble Sort}
%    \label{alg:examplse}
% \begin{algorithmic}
%    \STATE {\bfseries Input:} data $x_i$, size $m$
%    \REPEAT
%    \STATE Initialize $noChange = true$.
%    \FOR{$i=1$ {\bfseries to} $m-1$}
%    \IF{$x_i > x_{i+1}$}
%    \STATE Swap $x_i$ and $x_{i+1}$
%    \STATE $noChange = false$
%    \ENDIF
%    \ENDFOR
%    \UNTIL{$noChange$ is $true$}
% \end{algorithmic}
% \end{algorithm}

%我们提出了两种

\begin{figure}[tb]
\begin{center}
\centerline{\includegraphics[width=\columnwidth]{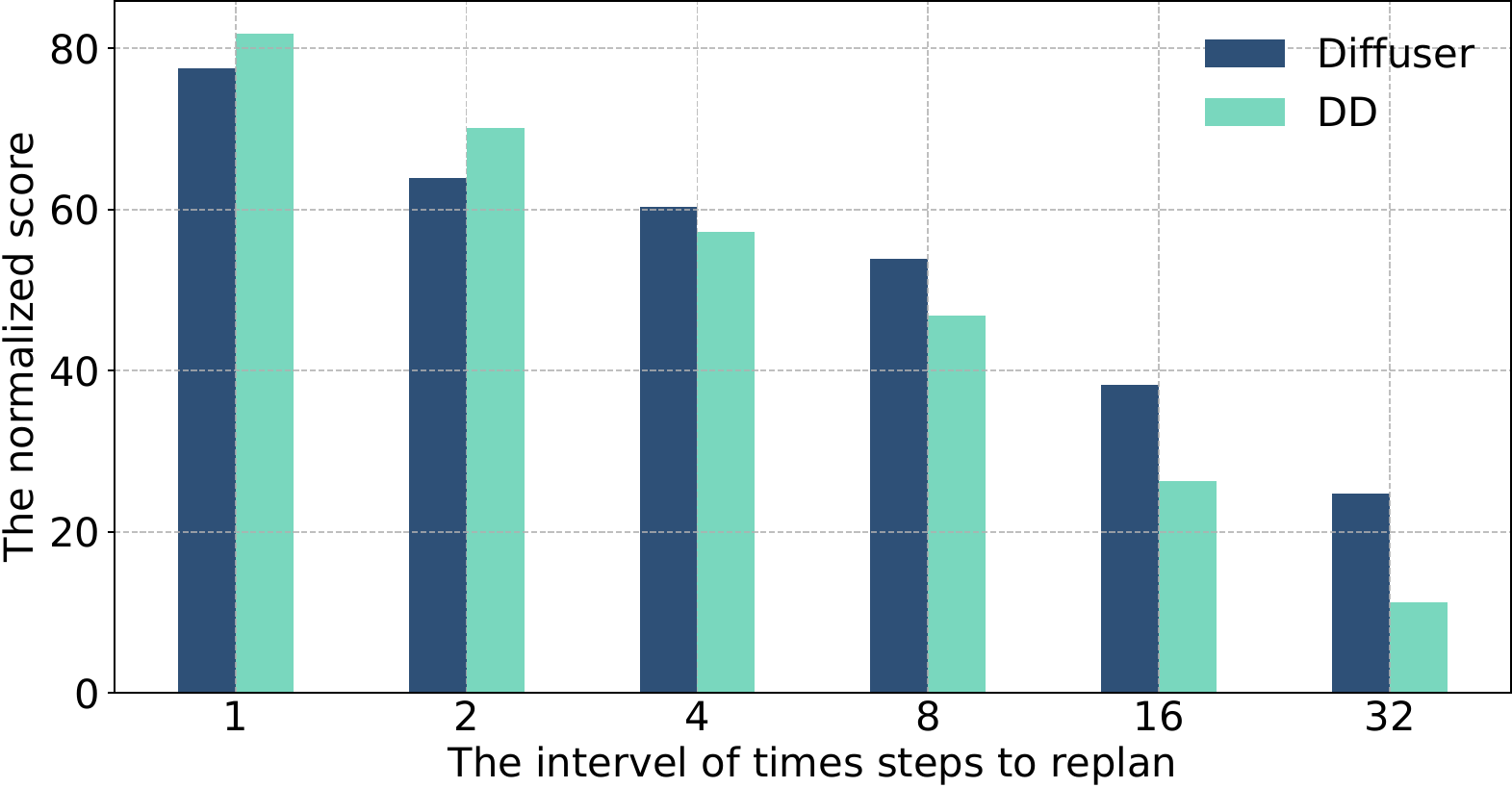}}
%\vspace{-5pt}
\caption{The average scores of previous diffusion planning methods on Mujoco when replan at different intervals.
}
\label{baseline}
\vspace{-10pt}
\end{center}
\end{figure}

\subsection{Acting with Inverse Dynamic Model}
%在前面的section中，为了更公平得与diffuser比较，我们使用了和diffuser相同的网络结构，并和diffuer一样生成完整的状态动作轨迹作为计划。但我们同样可以在计划时只产生状态，然后使用Inverse Dynamic Model获取动作：

%我们将这种变体称为
% In the previous sections, TDP generates trajectories with both states and actions as the plans and directly use the action from the plans for execution, 

In the previous sections, TDP leveraged the diffusion model to generate complete trajectories that included both states and actions for planning. Alternatively, the diffusion model can be used to generate trajectories consisting solely of states. Actions can then be inferred from consecutive states using an inverse dynamics model \cite{idm1,dd}: {\small$a_t=\mathcal{I}_\psi(s_t,s_{t+1}).$}
We refer to this variant of TDP as TDP\textsubscript{Inv}. To obtain actions through the inverse dynamics model, the next state {\small$s_{t+1}$} in the generated plan needs to undergo a complete denoising process. We adjust the Triangular Initial Planning and Plan Refinement processes such that the number of denoising steps is reduced by {\small$\kappa$} every two timesteps.
Thus, the candidate plans generated at each timestep can be represented as 
{\small$\{s_t,s_{t+1},s_{t+2}^{\kappa},s_{t+3}^{\kappa}\cdots,s_{t+H-2}^{K-\kappa},a_{t+H-1}^{K-\kappa}\}_M$}. 
During training, TDP\textsubscript{Inv} introduces an additional loss function to train the inverse dynamics model:
{\small$\mathcal{L}(\psi)=\mathbb{E}_{(s_t,a,s_{t+1}) \in D}[\Vert a_{t+1}- \mathcal{I}_\psi(s_t,s_{t+1})\Vert^2].$}
%We provide the details of other components that need to be modified in the Appendix.

\section{Experiment}
In this section, we explore the efficacy of the proposed TDP on various decision-making tasks. In particular, we aim to answer the following research questions:  \textbf{(1)} Compared with diffusion planning methods which generate new plans at each step, to what extent can TDP improve decision-making efficiency?  \textbf{(2)} How is the performance of TDP on offline reinforcement learning tasks? \textbf{(3)} What impact does the design choice of TDP have on performance?

\subsection{Experimental Setup}

\paragraph{Benchmarks.}
We evaluate the capacity of our method to recover an effective policy from previously collected data on the D4RL offline RL benchmark suite \cite{d4rl} which contains tasks of various domains. We use locomotion tasks of Gym-MuJoCo \cite{gym} which contains heterogeneous data of varying quality. We also use FrankaKitchen \cite{kitchen} tasks which require long-term credit assignment ability. 

\paragraph{Baselines.}
We compare our method with the basic imitation learning algorithm BC and existing offline RL methods, including model-free approaches like CQL \cite{cql} and IQL \cite{iql}, return conditioning approaches like Decision Transformer (DT) \cite{dt} and model-based approaches such as trajectory transformer (TT) \cite{tt} and MoReL \cite{morel}. For diffusion planning methods, we compare with Diffuser \cite{diffuser} and Decision Diffuser (DD) \cite{dd}. We also compare TDP with methods that enhance the efficiency of the Diffuser by reducing the number of diffusion steps, including Warm-Start (WS) \cite{diffuser}, Tree-Warm-Start (TWS) \cite{tree} and DDIM \cite{ddim}. We provide more details in the Appendix.

\begin{table}[tb]
\centering
{\small
\begin{tabular}{@{}cccc@{}}
\toprule
\textbf{Dataset}           & \textbf{TDP\textsubscript{All}} & \textbf{TDP\textsubscript{Best}} & \textbf{TDP}   \\ \midrule
Hopper-Medium-Expert & 39.6            & \textbf{111.5}   & \textbf{112.4} \\
Hopper-Medium     & 56.4            & \textbf{73.3}    & \textbf{74.0}  \\
   Hopper-Medium-Replay & 86.2            & \textbf{94.9}    & \textbf{94.9}  \\ \midrule
\textbf{Average}           & 60.73           & \textbf{93.2}    & \textbf{93.7}  \\ \bottomrule
\end{tabular}
}
\caption{Ablation results for candidate selection. TDP\textsubscript{All} and TDP\textsubscript{Best} means using all candidate plans or only using the best plan for planning refinement respectively.}
\label{abcs}
%\vspace{-10pt}
\end{table}

\begin{table}[t]
\setlength{\tabcolsep}{3mm}
\centering
{\small
\begin{tabular}{@{}ccccc@{}}
\toprule
\textbf{Dataset}     & \textbf{WS} & \textbf{TWS} & \textbf{DDIM} & \textbf{TDP}   \\ \midrule
Medium-Expert & 0.156               & 0.214                    & 0.379         & \textbf{0.049} \\
Medium        & 0.267               & 0.309                    & -             & \textbf{0.11}  \\
Medium-Replay & 0.28                & 0.298                    & 0.185         & \textbf{0.067} \\ \bottomrule
\end{tabular}%
}
\caption{Comparison in terms of runtime cost (s) with methods that enhance the efficiency of Diffuser by reducing the number of diffusion steps. }
\label{reduce}

\end{table}

\subsection{Results}
\paragraph{Efficiency.}
To evaluate the efficiency of the proposed Temporal Diffusion Planner (TDP), we measured the wall-clock runtime for a single decision-making step (one action inference) and calculated the corresponding decision frequency. The results are summarized in Table \ref{efficiency}.
For Gym-Mujoco tasks, TDP and TDP\textsubscript{Inv} reduced runtime costs to 11.7\% and 17\% of Diffuser, and to 5.2\% and 7.6\% of Decision Diffuser (DD), respectively. Similarly, for FrankaKitchen tasks, TDP and TDP\textsubscript{Inv} reduced runtime costs to 7.1\% and 8.6\% of Diffuser, and to 3.3\% and 4.0\% of DD, respectively.
On average, TDP increased decision frequency by 11 times and 24.8 times compared to Diffuser and DD, respectively, while TDP\textsubscript{Inv} achieved improvements of 8.6 times and 19.4 times.
These results highlight that TDP significantly enhances decision-making efficiency. 
This improvement is substantially due to the reduction in the average number of denoising steps required per decision-making step. Through temporal diffusion, we can mostly use plan refinement to make decisions instead of generating plans from scratch, reducing the diffusion steps, and thereby improve the efficiency. We record the average replanning interval of TDP and TDP\textsubscript{Inv}, which are 37.3 and 31.5 for Mujoco, 62.3 and 302.3 for Kitchen. Besides, as Figure \ref{baseline} shows, Diffuser and DD experience significant performance degradation without replanning at each step, so they cannot improve decision efficiency by reusing plans. For methods which enhance the efficiency of the Diffuser by reducing the number of diffusion steps, we evaluate them across varying denoising steps in the hopper environment. Table \ref{reduce} reports the minimal runtime of each method under the constraint that performance remains at or above 95\% of the Diffuser, denoting with a dash when performance preservation is unattainable with any tested denoising steps. The results demonstrate TDP's significant efficiency advantages over Warm-Start (WS), Tree-Warm-Start (TWS) and DDIM. %We provide the detailed results of these methods in the Appendix.

\paragraph{Performance.}
We evaluated TDP across different offline reinforcement learning tasks from the D4RL benchmark to verify whether it can maintain performance when improving the efficiency. We report the normalized scores of different algorithms in Table \ref{main_results}. We find that the proposed TDP achieves higher or comparable performance compared to baseline methods. For diffusion planning methods, Diffuser is the most important baseline as TDP uses the same diffusion model architecture and the same approach for conditional sampling.
The superior performance of TDP compared to Diffuser demonstrates that the proposed temporal diffusion approach not only improves decision efficiency but also enhances performance. Compared to Decision Diffuser, which employs a different conditional sampling method and requires longer decision times, TDP achieves comparable performance. Moreover, our variant TDP\textsubscript{Inv}
 , which predicts actions using an inverse dynamics model, slightly outperforms Decision Diffuser in average scores on D4RL locomotion tasks. On the D4RL Kitchen tasks, which require long-term credit assignment, TDP significantly outperforms Diffuser. This may indicate that refining and reusing previous plans to ensure action continuity is beneficial for long-horizon tasks. %We also provide experiments results of maze environment in the Appendix.

\subsection{Ablation Study}
\label{ablation}
\paragraph{Candidate Selection during Refinement.}
In TDP, plan refinement selects the top $m$
m plans from the candidate plans generated in the previous step based on the objective function $\mathcal{J}_{\phi}(\tau)$ for further refinement. Across all environments, we set the total number of candidate plans as 64 and $m$ as 8. To analyze the impact of candidate selection on performance, we conduct experiments in the Hopper environment and report the results in Table \ref{abcs}. We find that retaining all candidate plans led to a significant performance drop while keeping only the best plan resulted in only a slight performance decrease. This indicates that selecting only the better plans from the candidates is essential, but retaining multiple plans to ensure diversity is not critical.

\begin{figure}[t]
\centering
\centerline{\includegraphics[width=\columnwidth]{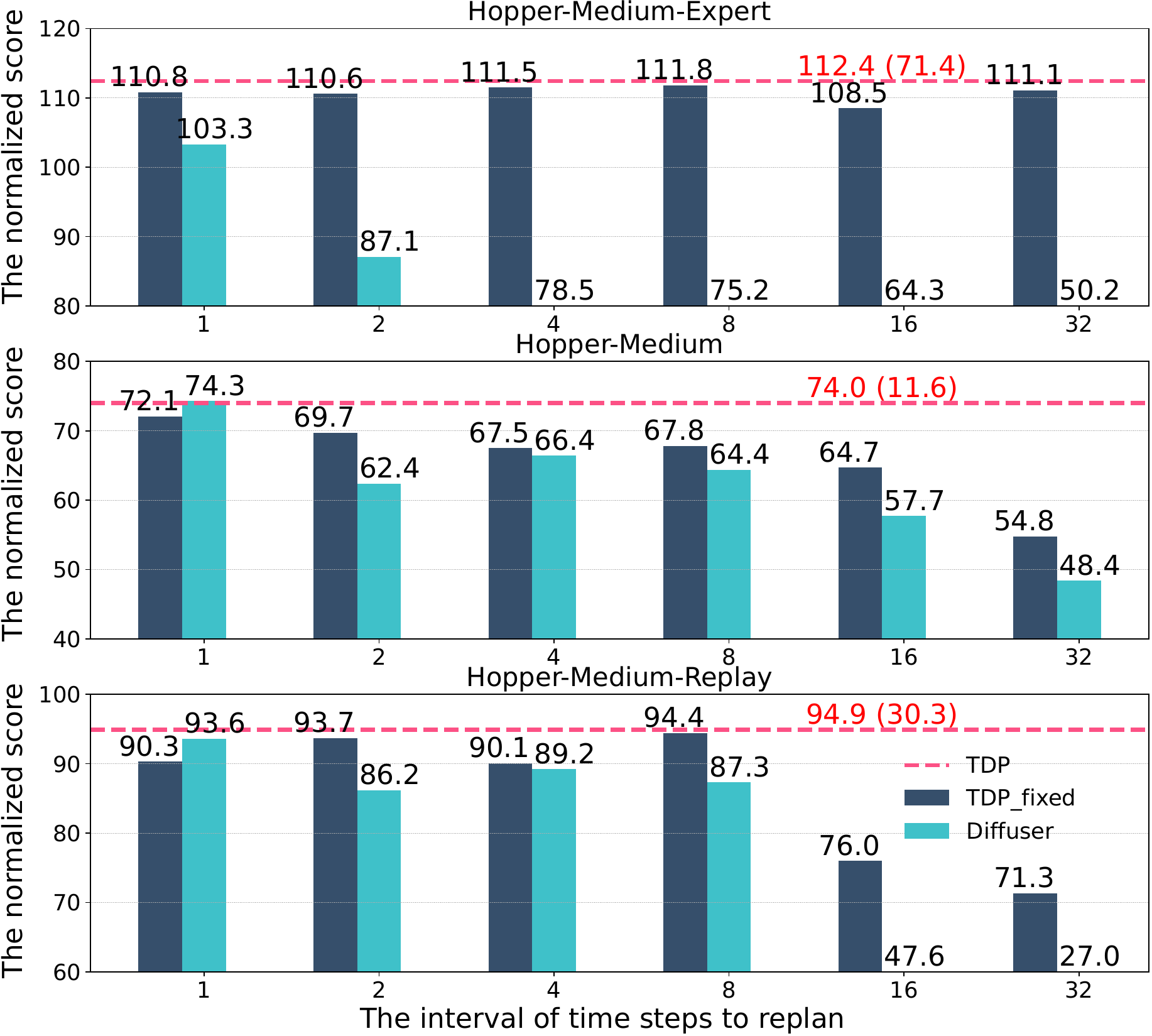}}
\caption{The normalized scores of TDP for replanning at different fixed intervals. We present the normalized scores when automatic replan with the dashed lines.
}
\label{fixinter}
%\vspace{-10pt}
\end{figure}

\begin{figure}[tb]
\centering
\centerline{\includegraphics[width=0.95\columnwidth]{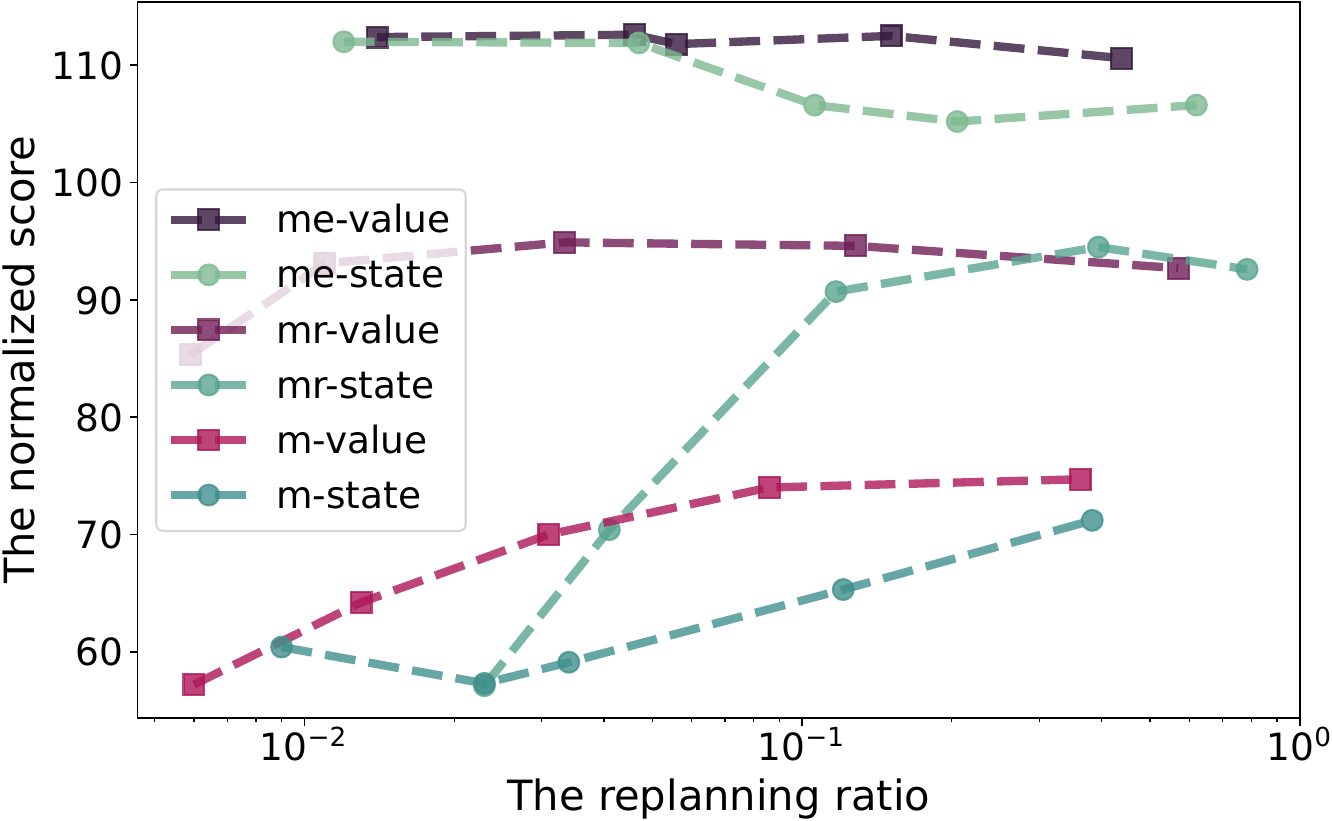}}
%\vspace{-5pt}
\caption{Ablation results for criteria of automatic replan. }
\label{replan}
%\vspace{-12pt}
\end{figure}
\paragraph{Replan at Fixed intervals.}

We verify the effect of the proposed automatic replanning mechanism by ablating it and generating plans from scratch at a fixed interval: $[1,2,4,8,16,32]$. We carry out experiments in the hopper environment and report the normalized scores in Figure \ref{fixinter}. For all data quality, TDP with the automated replanning mechanism received higher normalized scores, indicating that our automated scheduling mechanism can effectively determine when it is better to replan. Besides, for Medium-Expert data, replanning at different intervals has little effect on performance, while when using Medium and Medium-replay data, performance decreases significantly when the interval is too long. This shows that plan refinement cannot completely avoid failure when data quality is not good. We also observed that replanning each step did not always result in a better score, and when using Medium-replay data, fewer scheduled intervals resulted in a drop in performance, suggesting that the discontinuity of action caused by frequent plan switching has a negative impact on performance. We also compare the performance of the fixed-interval TDP with that of Diffuser. The main difference between them is whether using temporal diffusion. As the replanning interval increases, the performance of the Diffuser degrades much faster than TDP. This indicates that using temporal diffusion is critical for reusing the plan.

\paragraph{Replan with State-based criterion.}
We compare the effect of automatic replanning using state-based or value-based criterion through experiments in the Hopper environment. We use a series of thresholds for both criteria and record the corresponding performance and the ratio of replanning in all time steps. As Figure \ref{replan} shows, The performance of the value-based criterion is above the state-based criterion for all three quality of data. In addition, when using the value-based criterion, the performance gap corresponding to different replanning ratios is smaller. Therefore, we choose the value-based criterion for automatic replanning.

% \begin{table}[htb]
% \caption{\textbf{Ablation results for Replan Criterion.} TDP\textsubscript{State} means using the state-based criterion for automatic replanning in the temporal diffusion planner.}
% \centering
% \resizebox{0.7\columnwidth}{!}{%
% \begin{tabular}{@{}ccc@{}}
% \toprule
% \textbf{Dataset}           & \textbf{TDP\textsubscript{State}} & \textbf{TDP}   \\ \midrule
% Hopper-Medium-Expert & 112.0        & \textbf{112.4} \\
% Hopper-Medium     & 61.2         & \textbf{74.0}  \\
% Hopper-Medium-Replay & 77.2         & \textbf{94.9}  \\ \midrule
% \textbf{Average}           & 83.47        & \textbf{93.7}  \\ \bottomrule
% \end{tabular}
% }
% \end{table}

\section{Conclusion}
In this paper, we introduce the Temporal Diffusion Planner which improves the efficiency of diffusion planning. TDP uses a novel temporal diffusion approach to generate and improve plans over time and prevents performance degradation through automatic replanning. TDP achieves higher decision efficiency by reducing the average number of denoising steps required per step.  Experiments on D4RL show that, compared to previous works which generate new plans every time step, TDP achieves 11-24.8 times faster than previous mainstream methods and has higher or comparable performance. We focus on enhancing the basic diffusion planning while we will extent TDP to complex framework like multitask or hierarchical framework in the future work and hope TDP can inspire other works of diffusion planning.
\section*{Acknowledgements}
This work is partially supported by the Strategic Priority Research Program of the Chinese Academy of Sciences (Grants No.XDB0660200, XDB0660201, XDB0660202), the NSF of China (Grants No.62341411, 62222214), CAS Project for Young Scientists in Basic Research (YSBR-029) and Youth Innovation Promotion Association CAS.

\bibliography{anonymous-submission-latex-2026}

\end{document}